\documentclass{hld2025} % Author names withheld
\usepackage{hyperref}
\usepackage{booktabs}
\usepackage{natbib}
\usepackage{titlesec}
\usepackage{graphicx} 
\usepackage{caption}
\usepackage{subcaption}

\titlespacing*{\section}
  {0pt}{1.0ex plus .2ex minus .2ex}{1ex plus .2ex}
\titlespacing*{\subsection}
  {0pt}{0.8ex plus .2ex minus .2ex}{0.8ex plus .2ex}
\titlespacing*{\subsubsection}
  {0pt}{0.6ex plus .2ex minus .2ex}{0.5ex plus .2ex}

\title[Creativity and Attention ]{Origins of Creativity in Attention Based Diffusion Models}

\hldauthor{%
  \Name{Emma Finn}           \Email{efinn@college.harvard.edu}\\
  \Name{T. Anderson Keller}  \Email{t.anderson.keller@gmail.com}\\
  \Name{Manos Theodosis}     \Email{etheodosis@g.harvard.edu}\\
  \Name{Demba E. Ba}         \Email{demba@seas.harvard.edu}\\
  \addr{%
    CRISP Lab, Harvard John A.\ Paulson School of Engineering and Applied Sciences,\\
    and Kempner Institute for the Study of Natural and Artificial Intelligence,\\
    Harvard University, Cambridge, MA%
  }%
}

\begin{document}

\maketitle

\begin{abstract}
\looseness=-1
As diffusion models have become the tool of choice for image generation and as the quality of the images continues to improve, the question of how `creativity' originates in diffusion has become increasingly important. The score matching perspective on diffusion has proven particularly fruitful for understanding how and why diffusion models generate images that remain visually plausible while differing significantly from their training images. In particular, as explained in (Kamb \& Ganguli, 2024) and others, e.g., (Ambrogioni, 2023), theory suggests that if our score matching were optimal, we would only be able to recover training samples through our diffusion process. However, as shown by Kamb \& Ganguli, (2024), in diffusion models where the score is parametrized by a simple CNN, the inductive biases of the CNN itself (translation equivariance and locality) allow the model to generate samples that globally do not match any training samples, but are rather patch-wise `mosaics'. Despite the widespread use of UNet architectures with self‐attention as the score backbone in diffusion models, the theoretical role of attention in score networks remains largely unexplored. In this work, we take a preliminary step in this direction to extend this theory to the case of diffusion models whose score is parametrized by a CNN with a final self-attention layer. We show that our theory suggests that self-attention will induce a globally image-consistent arrangement of local features beyond the patch-level in generated samples, and we verify this behavior empirically on a carefully crafted dataset. 
\end{abstract}

%\begin{keywords}%
%  List of keywords%
%\end{keywords}

\section{Introduction}

Diffusion models have become the premier tool for image generation in the last decade \cite{sohl-dickstein2015deep}. Their capacity to generate visually plausible images that generalize beyond the training dataset makes them extremely useful, but this capacity is not well understood \cite{ho2020denoising}. Diffusion models operate by performing a series of transformations that map the underlying distribution of images to a centered, multivariate Gaussian and then learning the reverse process. One of the most fruitful approaches for understanding creativity in diffusion models has been the score matching perspective, where a relatively small neural network is trained to approximate the derivative of the log-likelihood of the underlying image distribution \cite{pml2Book}. However, a large body of work has demonstrated that if this score-approximation is exact, a diffusion model can only return training samples: it is not creative at all \cite{kamb2024analytict, li2024goodscoredoeslead}. Kamb and Ganguli in \cite{kamb2024analytict} offered an important first step in understanding why diffusion models are able to generalize extremely well despite theory suggesting that well-trained diffusion models should memorize \cite{biroli2024dynamicalregimes}. In particular, they provide a complete theory for diffusion models whose score approximator is a convolutional neural network (CNN). Because CNNs have two inductive biases, translational equivariance and locality, they solve analytically for a `score machine' with those two inductive biases. They demonstrated that for CNN-backed diffusion models, their theory ``partially predicts the results'' of pre-trained diffusion models \cite{kamb2024analytict}. In practice, however, state-of-the-art diffusion models use a much more complex score estimator network. In particular, most these models use a U-Net structure with self-attention blocks throughout \cite{sehwag2024minimaldiffusion, ho2020denoising}, though recent papers have explored a fully transformer-based score network \cite{peebles2023scalablediffusionmodelstransformers}. Both architectures violate the local assumption strongly and the translational equivariance assumption weakly. \\

\noindent In this work, we propose a theory for CNNs with a single self-attention layer at the very end, which provides a first step towards bridging the gap between the existing theory and state-of-the-art models. In particular, we derive a simple theoretical example that suggests that self-attention may play the role of enforcing global self-consistency in the other-wise local patch-mosaic construction of Kamb and Ganguli \cite{kamb2024analytict}. Empirically, we then validate this intuition on a simple toy dataset, showing that samples are far more globally self-consistent with attention than without. We propose this as a first step towards understanding the role of attention in the creativity of diffusion models.

% In particular, we demonstrate that the score matching framework can be extended to attention-based models and that our theory predicts the behavior of attention-based diffusion models.  

\section{Equivariant Score Machine with Attention}
To gain intuition for the form of the optimal score function with attention, we first will analyze a model with full attention over all patches. Then, to provide a tractable closed form solution, we will make the additional assumption that attention is `top-1', and show that this intuition holds.
\subsection{Simple CNN with Full Attention}
\label{Simple CNN with Attention}
\noindent We begin with the following notation, following \cite{kamb2024analytict}: let $\phi$ be an arbitrary image in the diffusion process, and for each pixel location $x \in \Lambda$, we write $\phi_x$ for the pixel value of $\phi$ at location $x$. Let $\Omega_x \subset \Lambda$ be the set of pixels in the patch centered at $x$. Let $\Phi$ be an arbitrary patch. Let $\pi_t$ be the distribution over (noisy) images at time $t$. The true score at $x$ is  \(s_t[\phi](x)\;=\;\nabla_{\phi_x}\log\pi_t(\phi)\,.\) \\

\noindent We embed each patch 
% via \( z_x \;=\;g(\phi_{\Omega_x})\;\in\; \mathbb R^d\)
via \(g(\phi_{\Omega_x})\;\in\; \mathbb R^d\) 
where $g$ is a convolutional embedding network, which can be thought of as the first portion of our score-approximation network. Our learnable parameters are in $g$. We then define the full score function estimator $\tilde{g}[\phi](x)$, which corresponds to a single layer of attention on top of a CNN embedding.
% Differing from \cite{kamb2024analytict}, in this work, instead of a purely local estimator $g(\phi_{\Omega_x})$, we define:
% \[
% \tilde g[\phi](x)
% \;=\;
% z_x \;+\;
% \sum_{y}\alpha_{xy}\,z_y
% \qquad
% \alpha_{xy}
% \;=\;
% \frac{\exp\!\bigl(\langle z_x,z_y\rangle\bigr)}
%      {\sum_{y'}\exp\!\bigl(\langle z_x,z_{y'}\rangle\bigr)}
% \]
\begin{equation}
\tilde g[\phi](x)
\;=\;
g(\phi_{\Omega_x}) \;+\;
\sum_{y}\alpha_{xy}\,g(\phi_{\Omega_y}), 
\quad \text{where} \quad 
\alpha_{xy}
\;=\;
\frac{\exp\!\bigl(\langle g(\phi_{\Omega_x}),g(\phi_{\Omega_y})\rangle\bigr)}
     {\sum_{y'}\exp\!\bigl(\langle g(\phi_{\Omega_x}),g(\phi_{\Omega_{y'}})\rangle\bigr)},
\end{equation}
so that each location $x$ `attends' to all patches in the image, where the sum over $y$ runs over all other patch centers $y \in \Lambda$ in the image $\phi$. In particular, we apply the simplest kind of attention over top of a convolutional neural network. The standard attention framework involves three learnable projection matrices $W_Q, W_K$ and $W_V$ that embeds our input set of tokens, $X$ into queries $Q$, keys, $K$, and values, $V$. Attention weights are then given by  $\text{Attention(Q,K,V)} = \text{softmax}\left(\frac{Q K^T}{\sqrt{d}}\right) V$. In our case, we downgrade $W_Q$, $W_K$, and $W_V$ from learnable parameters into identity matrices, and recover the form 
\begin{align}
    \text{Attention(Q,K,V)} = \text{softmax}\left(\frac{XX^T}{\sqrt{d}}\right) V
\end{align}
 which gives us exact the entry-wise formulation above, which says that $\alpha_{xy} = \frac{\exp(x_i x_j)}{\sum_{k=1}^n \exp(x_i \cdot x_k)}$. We drop the scalar $\frac{1}{\sqrt{d}}$ for simplicity here.
 
\noindent Adopting this functional form, the corresponding score matching loss becomes
\[
\mathcal L
\;=\;
\sum_{x}\mathbb{E}_{\phi\sim\pi_t}
\bigl\|\tilde g[\phi](x)\;-\;s_t[\phi](x)\bigr\|^2.
\]
We mirror the derivation in \cite{kamb2024analytict} and write this expectation as an integral over $\pi_t$ and impose the stationary condition w.r.t. $g$ (since the attention layer has no parameters), yielding:
\begin{align}
 %    0 = \frac{\delta\mathcal L}{\delta g(\Phi)} \;=  \sum_x\frac{\delta}{\delta g(\Phi)} \int \pi_t(\phi)  [||g(\phi_{\Omega_x}) + \sum_y(\text{softmax}_y(g(\phi_{\Omega_x}), g(\phi_{\Omega_y}))g(\phi_{\Omega_y}) - s[\phi](x)||^2]  d \phi
 % \quad\forall\,\Phi
     0 = \frac{\delta\mathcal L}{\delta g(\Phi)} \;=  \sum_x\frac{\delta}{\delta g(\Phi)} \int \pi_t(\phi)  [||\tilde{g}[\phi](x) - s[\phi](x)||^2]  d \phi
 \quad\forall\,\Phi.
\end{align}

\noindent See Appendix A for details on the derivation.  Eventually we find that, setting $\mu_x := \sum_{z} \alpha_{xz}(\phi)\,g(\phi_{\Omega_{z}})$:

\begin{align}
0 =\ & 2 \sum_x \int \pi_t [ g(\phi_x) + \sum_y \alpha_{xy} g(\phi_{\Omega_y}) - s[\phi](x) ]^T 
\delta(\phi_{\Omega_x} - \Phi)\ d\phi \nonumber \\
& + \int \pi_t [ g(\phi_x) + \sum_y \alpha_{xy} g(\phi_{\Omega_y}) - s[\phi](x) ]^T 
\sum_y \alpha_{xy} \left( I + (g(\phi_{\Omega_x}) - \mu_x) \delta(\phi_{\Omega_y} - \Phi) \right) d\phi
\label{keyeqn}
% 0 =\ & 2 \sum_x \int \pi_t \left[ \tilde{g}[\phi](x) - s[\phi](x) \right]^T 
% \delta(\phi_{\Omega_x} - \Phi)\ d\phi \nonumber \\
% & + \int \pi_t \left[ \tilde{g}[\phi](x) - s[\phi](x) \right]^T 
% \sum_y \alpha_{xy} \left( I + (g(\phi_{\Omega_x}) - \mu_x) \delta(\phi_{\Omega_y} - \Phi) \right) d\phi
\end{align}
\noindent Intuitively, the first term in our sum corresponds to a when the patch of interest $\Phi$ is the query, and the second term corresponds to when the patch is the key or the value. In particular, the $\delta(\phi_{\Omega_x} - \Phi)$ term matches exactly the CNN-case, and so encourages $g(\phi_{\Omega_x})$ to match the true score of $\phi_{\Omega_x}$, if it were considered a purely local, de-contextualized patch. The second term corresponds to the sum of gradients from every other patch $\phi_{\Omega_y}$ that attended to $x$, which ``encourages'' $g(\phi_{\Omega_x})$ to provide more information about the image at position $y$. If the weight $\alpha_{xy}$ is large, then, we should see that if there's error when we evaluate at position $x$, our gradient should push us to change the value of $g(\phi_{\Omega_y})$ so that it becomes even more useful in reconstructing the image at position $x$. Thus, our score function is a weighted average of the CNN score function evaluated at that patch's location and the score function evaluated at every other patch that is informative about the patch at position $x$. This will encourage ``copy-and-paste'' behavior of patches that often appear together in a given image since the score will move the image in the direction of self-consistency because of the second term. While the general functional‐gradient expression offers strong intuition for our empirical findings, deriving a closed‐form solution in full generality proves intractable. We therefore specialize to an informative, tractable case in the next section that partially matches our experimental setup and provides more direct insight into the behavior of the score machine.

\subsection{Simple CNN with Top-1 Attention}
For simplicity, we assume that our attention is a ``winner-take-all'' regime, meaning that only the most attended to patch contributes to the sum. In particular, we have 
\begin{equation}
\label{eqn:top1_attn}
      \sum_{y}\alpha_{xy}\,g(\phi_{\Omega_y})
      \;\longrightarrow\;
      g\bigl(\phi_{\Omega_{y^*(x)}}\bigr),
      \quad
      y^*(x)=\arg\max_y\langle g(\phi_{\Omega_x}),\,g(\phi_{\Omega_y})\rangle.
\end{equation}

\noindent We also assume ``patch-independence'' under the distribution $\pi_t$ for all $t$, so that conditioning on $\phi_{\Omega_x} = \Phi$ does not change the distribution of $\phi_{\Omega_{y}}$ for $y \neq x$ and that our embedding $g$ is mean‐centered over patches (ie, \(\mathbb{E}_{\phi\sim\pi_t}\bigl[g(\phi_{\Omega_x})\bigr]=0 \quad\forall\,x\)).  We then substitute the top-1 attention form of Equation \ref{eqn:top1_attn} into Equation \ref{keyeqn}. When we expand the the second term in Equation \ref{keyeqn}, we find that since we've assumed that $g$ is mean-centered and that we have approximate patch independence, the second term goes away, leaving 

\begin{equation}
\begin{aligned}
% 0=\;&
% \sum_{x}\!\int\!\pi_t(\phi)\;
%    2\bigl(g_x+g_{y^{*}(x)}-s[\phi](x)\bigr)^{\!\top}
%    \delta(\phi_{\Omega_x}-\Phi)\,d\phi \\[4pt]
% &+\;\sum_{x}\!\int\!\pi_t(\phi)\;
%    2\bigl(g_x+g_{y^{*}(x)}-s[\phi](x)\bigr)^{\!\top}
%    \delta(\phi_{\Omega_{y^{*}(x)}}-\Phi)\,d\phi .
0=\;&
\sum_{x}\!\int\!\pi_t(\phi)\;
   2\bigl(\tilde{g}[\phi](x)-s[\phi](x)\bigr)^{\!\top}
   \delta(\phi_{\Omega_x}-\Phi)\,d\phi \\[4pt]
&+\;\sum_{x}\!\int\!\pi_t(\phi)\;
   2\bigl(\tilde{g}[\phi](x)-s[\phi](x)\bigr)^{\!\top}
   \delta(\phi_{\Omega_{y^{*}(x)}}-\Phi)\,d\phi .
\end{aligned}
\end{equation}
Since there is a deterministic mapping between a given patch $\phi_{\Omega_x}$ and its most attended patch $\phi_{\Omega_{y^*(x)}}$, we see that the integrals with the delta peaks give closed form solutions:
\begin{equation}
    \int \tilde{g}[\phi](x)\delta(\phi_{\Omega_x} - \Phi) d\phi = \int \tilde{g}[\phi](x)\delta(\phi_{\Omega_y^*(x)} - \Phi) d\phi = g[\Phi] + g[\Phi^*] := \tilde{g}[\Phi],
\end{equation}
where $g[\Phi^*]$ denotes the output of the model on the patch most attended to by $\Phi$.
% 
% 
% from which, we aim to solve for $\tilde{g}(\Phi)$.
% 
% Assuming that the patches cannot attend to themselves, i.e. $y^*(x) \neq x$, we get the following two equalities: 
% \begin{equation}
%     \int g(\phi_{\Omega_{y^*(x)}})\delta(\phi_{\Omega_x} - \Phi) d\phi = 0  \quad \& \quad \int g(\phi_{\Omega_{x}})\delta(\phi_{\Omega_y^*(x)} - \Phi) d\phi = 0
% \end{equation}
% Potentially this is 0 because we have that $\int g(\phi_{\Omega_{y^*(x)}})\delta(\phi_{\Omega_x} - \Phi) d\phi ?= \pi_t (\phi_{\Omega_x} = \Phi) \mathbb{E}[g(\phi_{\Omega_{y^*(x)}} | \phi_{\Omega_x} =\Phi]$
% 
% $\int g(\phi_{\Omega_y}) \delta(\phi_{\Omega_x} - \Phi) \pi_t(\phi) d\phi = \int g(\omega_y) \pi_t( \omega_{y(x)}, \omega_x = \Phi) d\phi = \int g(\omega_{y(x)}) \pi_t( \omega_{y(x)}= \Phi) \pi_t (\omega_x = \Phi) d\phi  =  \pi_t (\omega_x = \Phi) \int g(\omega_{y(x)}) \pi_t( \omega_{y(x)}= \Phi)d\phi $
% 
% Distributing the deltas, and then removing these terms, we get:
% \begin{equation}
% \begin{aligned}
% 0=\;&
% \sum_{x}\!\int\!\bigl(\pi_t(\phi)\;g(\phi_{\Omega_x})\delta(\phi_{\Omega_x}-\Phi)-\pi_t(\phi)s[\phi](x)
%    \delta(\phi_{\Omega_x}-\Phi)\bigr)\,d\phi \\[4pt]
% &+\;
% \sum_{x}\!\int\!\bigl(\pi_t(\phi)\;g(\phi_{\Omega_{y^{*}(x)}})\delta(\phi_{\Omega_{y^{*}(x)}}-\Phi)-\pi_t(\phi)s[\phi](x)
%    \delta(\phi_{\Omega_{y^{*}(x)}}-\Phi)\bigr)\,d\phi  
% \end{aligned}
% \end{equation}
% 
Distributing the deltas, integrating, and eventually dividing out by $\bigl[\pi_t(\phi_{\Omega_x} = \Phi) + \pi_t(\phi_{\Omega_{y^*(x)}} 
= \Phi)\bigr]$, we arrive at the following solution for the optimal score function $\tilde{g}[\Phi]$:

%Thus, distributing the deltas, integrating over these terms, and moving them to the other side:
%\begin{equation}
%\begin{aligned}
%& \tilde{g}(\Phi)\sum_{x}\pi_t(\phi_{\Omega_x} = \Phi) + \tilde{g}(\Phi)\sum_{x}\pi_t(\phi_{\Omega_{y^*(x)}} 
%= \Phi) = \\ &\;
%\sum_{x}\!\int\!\pi_t(\phi)s[\phi](x)
 %  \delta(\phi_{\Omega_x}-\Phi)\,d\phi
%+\;
%\sum_{x}\!\int\!\pi_t(\phi)s[\phi](x)
%   \delta(\phi_{\Omega_{y^{*}(x)}}-\Phi)\,d\phi
%\end{aligned}
%\end{equation}
%By the linearity of the integrals and sums we get:
%\begin{equation}
%\begin{aligned}
 %\tilde{g}(\Phi)\sum_{x}\bigl[\pi_t(\phi_{\Omega_x} = \Phi) + \pi_t(\phi_{\Omega_{y^*(x)}} 
%= \Phi)\bigr]  % & = \;
%\sum_{x}\!\int\!\pi_t(\phi)s[\phi](x)\bigl[
   %\delta(\phi_{\Omega_x}-\Phi) +
   %\delta(\phi_{\Omega_{y^{*}(x)}}-\Phi)\bigr]\,d\phi \\
   %& = \sum_{x}\!\int\!\nabla_{\phi(x)}\pi_t(\phi)\bigl[
   %\delta(\phi_{\Omega_x}-\Phi) +
   %\delta(\phi_{\Omega_{y^{*}(x)}}-\Phi)\bigr]\,d\phi \\
   %& = \sum_{x}\!\nabla_{\Phi(0)}\bigl[\pi_t(\phi_{\Omega_x} = \Phi ) + \pi_t(\phi_{\Omega_{y^{*}%(x)}}=\Phi)\bigr]
%\end{aligned}
%\end{equation}
%Dividing by $\bigl[\pi_t(\phi_{\Omega_x} = \Phi) + \pi_t(\phi_{\Omega_{y^*(x)}} 
%= \Phi)\bigr]$:
\begin{equation}
\begin{aligned}
 \tilde{g}[\Phi] = \nabla_{\Phi(0)}\log\sum_{x}\!\bigl[\pi_t(\phi_{\Omega_x} = \Phi ) + \pi_t(\phi_{\Omega_{y^{*}(x)}}=\Phi)\bigr]
 \end{aligned}
 \end{equation}
 \looseness=-1
We refer readers to the appendix for the full derivation. 
If we then continue with the same procedure as equations 36-40 of \cite{kamb2024analytict} we see that the optimal score function is the gradient of a mixture of Gaussians centered at patches in the dataset \emph{combined} with a mixture of Gaussians centered at the most attended corresponding patches. Intuitively, this implies that during the diffusion reverse process, each patch will be pulled towards its corresponding closest patch from the dataset, \emph{but also} the corresponding closest patch \emph{from the current image}. This second term is exactly what we term global self-consistency of the generated patches -- they will tend to align with other patches in a current image. 

\section{Experiments}
We validate our theory by showing that a simple attention-based diffusion model can learn and effectively reproduce such self-consistent structures in images while a CNN-based diffusion model struggles. We evaluate the capacity to construct consistent images by measuring how often the samples generated by our diffusion model obey the rules of this dataset. First, we construct a dataset to test this consistency property. Our dataset consists of 2048 4x4 RBG images in which each 2×2 block is filled with one of three possible color–color pairings (red/green vs. yellow/blue; red/yellow vs. green/blue; red/blue vs. green/yellow). These parings are randomized across images but are consistent within images. To construct our dataset, we randomly select one of three pattern types for each image (two disjoint color pairings, for example, red and green, and blue and yellow). Then, for each of the four quadrants in our image, we randomly select either a block of the first pairing type or a block of the second pairing type. In particular, this means that to successfully generate samples, our score must be sensitive to these consistent ``key-value'' pairings \emph{within} images rather than the potentially inconsistent pairings in the dataset as a whole.\\

\noindent We train four simple diffusion models on this dataset: a pure CNN-based model, a CNN-based model with top-1 attention, a CNN-based model with attention with identity key and query matrices, and a CNN + Attention model. We use a standard DDPM setup and implement our score network (or, equivalently, noise predictor) as a very simple 2 layer CNN with 2 convolutional layers, where the first convolutional layer is a 2x2 convolution with stride 2 and hidden dimension 32, and our second is a transpose convolution also with a 2x2 kernel and stride length of 2. This kernel and stride length help to encode the desired inductive biases and ensures that keys are encoded with their associated values. Our CNN+Attention model has exactly the same CNN base but includes a single-headed self-attention block with learnable Q, K, V matrices right before the final projection. We train all models identically. Details on the training and more information about the two modified attention models and their performance can be found in Appendix B. \\ 

\begin{figure}[t]
  \centering

  % Image
  \begin{minipage}[t]{1.0\textwidth}
    \centering
    \includegraphics[width=\linewidth]{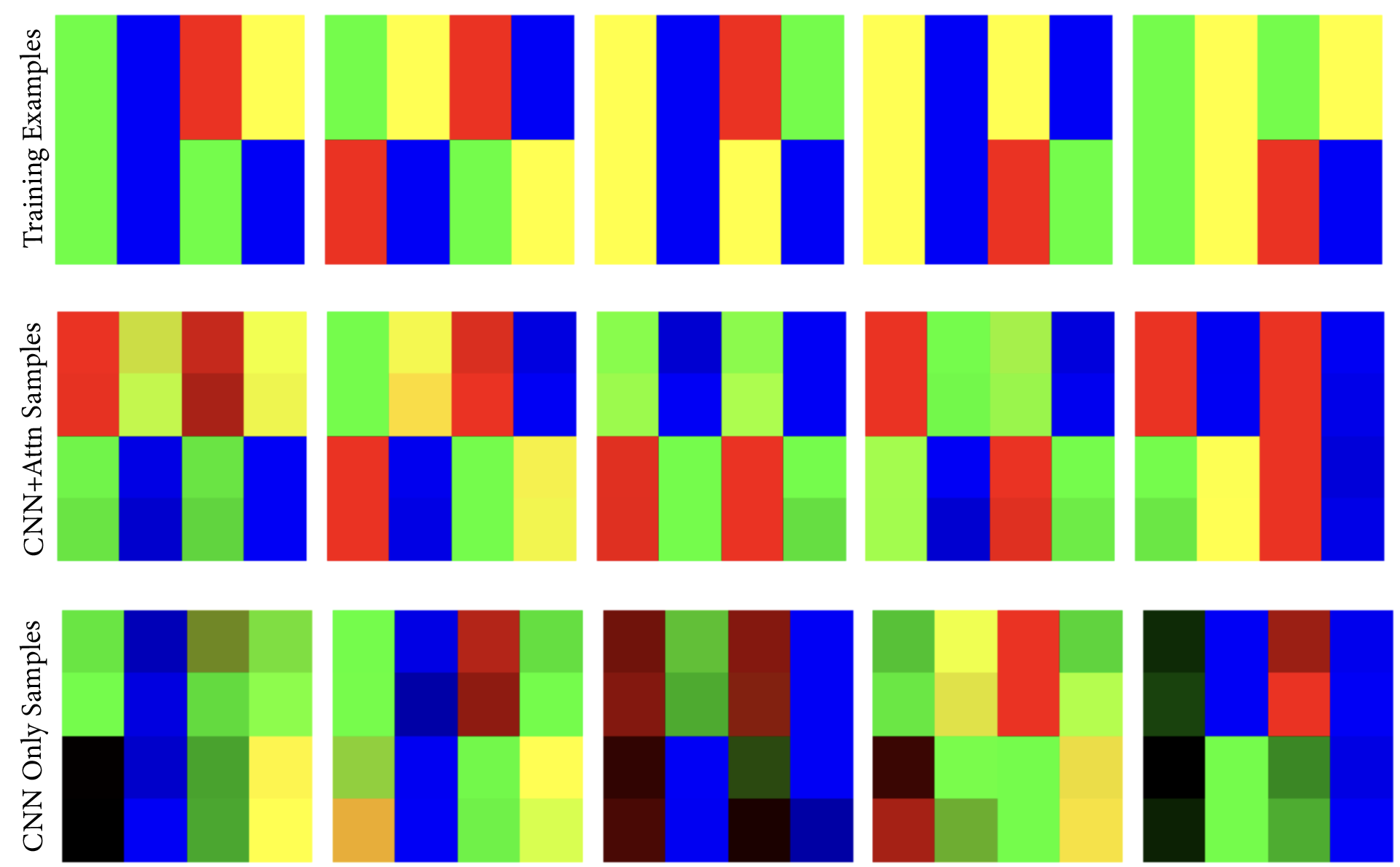}
    \captionof{figure}{Dataset and generated images.}
    \label{fig:generated_samples}
  \end{minipage}

  \vspace{1em}

  % Table
  \begin{minipage}[t]{0.5\textwidth}
    \centering
    \begin{tabular}{lc}
      \toprule
      \textbf{Model}    & \textbf{Consistency} \\
      \midrule
      Random Baseline   & 5.38\%               \\
      CNN               & 10.88\%              \\
      CNN + Attention   & 64.03\%              \\
      \bottomrule
    \end{tabular}
    \captionof{table}{Self‐consistency of generated samples.}
    \label{tab:consistency}
  \end{minipage}
\end{figure}

\noindent Both qualitatively and quantitatively our attention-based model demonstrated increased consistency within images and higher quality image generation overall, as shown in Figure~\ref{fig:generated_samples} and Table \ref{tab:consistency}. It is clear that the CNN+Attention samples are both more visually consistent with the training samples and reproduce blocks of the correct size and colors, showing that attention is capturing the consistency of ``key-value'' pairs of a particular image. For each of our models (CNN and CNN+Attention) we generated 10k samples in 100 different runs. Over these 1 million samples generated by our attention-based model, $64.03\%$ were consistent. We say that an image is ``consistent'' if its mapping has key-value pairs that remain constant throughout the image. For example, if one quadrant of the image contains a green-yellow pair, then if green appears as a key in any other quadrant it must be followed by yellow. We require this for each paring that appears in the image. In practice, for a given generated image, we map each pixel to its nearest color. Then, we split the split the map of labels into four quadrants and checked whether those mappings were one of the permissible sets. Finally, we computed a robust  baseline ``consistency percentage'' by checking the consistency of 10k images generated by sampling four color pairs (with replacement), randomly assigning them to the four quadrants. We took the average percent consistent across 100 trials. \\

\noindent These consistency results match what our theory suggests: the second term in Equation~\ref{keyeqn} encourages the reconstruction of these key-value pairs that appear within a single image (self-consistency). In the CNN results, it is clear that while the model is reproducing features that appear across the dataset, including the quadrant layout, and some color pairing, it is unable to accurately construct these image-specific patterns without any non-local mechanism. This helps to explain some of the challenges that \cite{kamb2024analytict} faced when using their equivariant, local score machine to replicate the results of UNet, self-attention-based diffusion. 

\section{Discussion}
\noindent The above theory offers a promising new step towards a more theoretically grounded understanding of how creativity emerges in diffusion models beyond the simple, purely convolutional case. Our empirical results on a purpose‐built relational dataset support the theory: the CNN+Attention model generates significantly more self-consistent samples than its CNN-only counterpart. These findings suggest that even a single attention block can bridge the gap between local patch mosaics and full image consistency. While our theory here has been restricted to the very simple CNN+attention case, we hope to extend the theory in future work to a Unet + self-attention framework and replicate our results on larger datasets of natural images. 

\section*{Acknowledgments}
This work was supported by the Kempner Institute for the Study of Natural and Artificial Intelligence at Harvard University. We thank the Kempner for access to compute resources. We are also grateful to the CRISP group at Harvard SEAS for many thoughtful conversations. EF thanks the Calvin Coolidge Presidential Foundation for support during her undergraduate studies. 
This work is supported in part by the National Science Foundation under Cooperative Agreement PHY-2019786 (\href{ http://iaifi.org/}{The NSF AI Institute for Artificial Intelligence and Fundamental Interactions}).

% \begin{figure}[t]
%   \centering
%   \begin{minipage}[t]{0.50\textwidth}
%     \vspace{0pt}                      %
%     \includegraphics[width=\linewidth]{GeneratedSamples}
%   \end{minipage}\hfill
%   \begin{minipage}[t]{0.4\textwidth}
%     \vspace{0pt}                     
%     \centering
%     \captionof{table}{Self-consistency of Generated Samples}
%     \label{tab:summary}
%     \begin{tabular}{lc}
%       \toprule
%       Model     &  Consistency (\%) \\
%       \midrule
%       Random Baseline    & $5.38$ \%  \\
%       CNN    & 10.88 \%  \\
%       CNN + Attn & 64.03 \%   \\
%       \bottomrule
%     \end{tabular}
%   \end{minipage}

%   \caption{Left: dataset \& generated images. \quad Right: consistency scores across models.}
%   \label{fig:combined}
% \end{figure}

%\begin{figure}[t]
 % \centering
  % First figure: the image
  %\begin{minipage}[t]{0.75\textwidth}
   % \vspace{0pt}
  %  \centering
   % \includegraphics[width=\linewidth]{GeneratedSamples}
   % \caption{Dataset and generated images.}
   % \label{fig:generated_samples}
  %\end{minipage}\hfill
  % Second figure: the table
 % \begin{minipage}[t]{0.2 \textwidth}
  %  \vspace{10mm}
  %  \centering
    %\begin{tabular}{lc}
     % \toprule
     % \textbf{Model}     & \textbf{Consistency} \\
     % \midrule
     % Random Baseline    & 5.38\%  \\
     % CNN                & 10.88\% \\
     % CNN + Attention         & 64.03\% \\
     % \bottomrule
   % \end{tabular}
   % \captionof{table}{Self-consistency of generated samples.}
   % \label{tab:consistency}
  %\end{minipage}
%\end{figure}

%\begin{figure}
    %\centering
   % \includegraphics[width=0.7\linewidth]{GeneratedSamples}
   % \caption{Dataset and Generated Images}
   % \label{fig:examplepics}
%\end{figure}

\bibliographystyle{plainnat}   % or plain, abbrvnat, unsrt, etc.
\bibliography{ref}
\newpage
\clearpage
\appendix
\begin{appendix}
\section{Derivations}
\subsection{General Attention Derivation}
Here, we'll derive the self-consistency constraint for a simple attention mechanism mentioned in Section~\ref{Simple CNN with Attention}. We'll recall the notation used in \cite{kamb2024analytict}
\begin{enumerate}
    \item $\phi$ is an arbitrary image in the diffusion process
    \item $x$ is a pixel location and $\Lambda$ is the set of pixel locations in a given image.
    \item $\Omega_x$ is the neighborhood of pixels centered at x
    \item $\phi_{\Omega_x}$ is the patch of the image $\phi$ centered at $x$.
    \item $\pi_t$ is the underlying probability distribution of images at time $t$ 
    \item $s_t[\phi](x)$ is the true score at time t evaluated at position $x$ in the image $\phi$. 
\end{enumerate}

The original score matching loss function given in the paper \cite{kamb2024analytict} is given by
\begin{equation}
  \mathcal{L}
  = \sum_x
    \mathbb{E}_{\phi\sim\pi_t}
    \Bigl[ ||f(\phi_{\Omega_x})
        - s_t[\phi](x)
      \bigr\rVert^2
    \Bigr].
\end{equation}

\noindent We want to modify this so that we actually have $f(\phi_{\Omega_x})$ as an attention based update rather than just looking at one neighborhood of the image. First, we embed each patch using $g$, a convolutional embedding network, which can be thought of as the first portion of our score-approximation network. It is the only portion of our architecture with learnable parameters. 
\begin{enumerate}
    \item Let $z_x$ denote the embedding of the patch $\phi_{\Omega_x}$ in $\mathbb{R}^d$, where $z_x = g(\phi_{\Omega_x})$
    \item Let $(z_y)_{y \in \Lambda}$ be the collection of flattened patches of the image $\phi$ that we're currently looking at.
\end{enumerate}

Differing from \cite{kamb2024analytict}, in this work, instead of a purely local estimator $g(\phi_{\Omega_x})$, we define:
\[
 \tilde g[\phi](x)
\;=\;
 z_x \;+\;
 \sum_{y}\alpha_{xy}\,z_y
\qquad
 \alpha_{xy}
\;=\;
\frac{\exp\!\bigl(\langle z_x,z_y\rangle\bigr)}
      {\sum_{y'}\exp\!\bigl(\langle z_x,z_{y'}\rangle\bigr)}
\]

Now we define our new model, which we'll call $\tilde g$. 

\begin{equation}
    \tilde g(\phi)(x) = z_x + \sum_y \text{softmax}_y( \langle z_x, z_y \rangle) 
\end{equation}
where $\langle z_x, z_y \rangle$ denotes the dot product of our vector embeddings. We write $\alpha_{xy} = \text{softmax}_y( \langle z_x, z_y \rangle)$. The $y$ subscript says that we're normalizing over the y indexes so that for a fixed x, the sum of the $\alpha_{xy}$ is 1. Intuitively, a big value $\alpha_{xy}$ says that the position x should pay a lot of attention to the position y.

Thus, our loss function becomes

\begin{equation}
    \mathcal{L} = \sum_x \mathbb{E}_{\phi \sim \pi_t}[||z_x + \sum_y(\text{softmax}_y(z_x, z_y)z_y - s[\phi](x)||^2]
\end{equation}

Rewriting this in terms of our function g, we have
\begin{equation}
    \mathcal{L} = \sum_x \mathbb{E}_{\phi \sim \pi_t}[||g(\phi_{\Omega_x}) + \sum_y(\text{softmax}_y(g(\phi_{\Omega_x}), g(\phi_{\Omega_y}))g(\phi_{\Omega_y}) - s[\phi](x)||^2]
\end{equation}
Now we need to take the functional derivative of this. First we'll rewrite the expectation as an integral and then 
\begin{equation}
    \mathcal{L} = \sum_x \int[||g(\phi_{\Omega_x}) + \sum_y(\text{softmax}_y(g(\phi_{\Omega_x}), g(\phi_{\Omega_y}))g(\phi_{\Omega_y}) - s[\phi](x)||^2] \pi_t(\phi) d \phi
\end{equation}

Now we  find the function g which minimizes this loss. In particular, we'll assert 
\begin{equation}
    \frac{\delta \mathcal{L}}{\delta g(\Phi)} = 0 \text{ for each possible patch  } \Phi
\end{equation}

Now we need to take this derivative.

\begin{equation}
    0 =  \sum_x\frac{\delta}{\delta g(\Phi)} \int \pi_t(\phi)  [||g(\phi_{\Omega_x}) + \sum_y(\text{softmax}_y(g(\phi_{\Omega_x}), g(\phi_{\Omega_y}))g(\phi_{\Omega_y}) - s[\phi](x)||^2]  d \phi
\end{equation}

We can differentiate under the integral sign since the bounds don't depend on what we're taking the derivative with respect to, so 
\begin{equation}
    0 =  \sum_x \int \pi_t(\phi) \frac{\delta}{\delta g(\Phi)} \left( [||g(\phi_{\Omega_x}) + \sum_y(\text{softmax}_y(g(\phi_{\Omega_x}), g(\phi_{\Omega_y}))g(\phi_{\Omega_y}) - s[\phi](x)||^2] \right) d \phi
\end{equation}

Now we'll apply the vector-calculus chain rule to see that we get 
\begin{align}
    0 = \sum_x \int \pi_t(\phi)\, 2 \cdot 
    \Bigg(
        & g(\phi_{\Omega_x}) + 
        \sum_y \left( 
            \text{softmax}_y\big(g(\phi_{\Omega_x}), g(\phi_{\Omega_y})\big) 
            g(\phi_{\Omega_y}) 
        \right) \nonumber \\
        & - s[\phi](x)
    \Bigg)^{\!T}
    \frac{\delta}{\delta g(\Phi)}
    \Bigg(
        g(\phi_{\Omega_x}) + 
        \sum_y \left( 
            \text{softmax}_y\big(g(\phi_{\Omega_x}), g(\phi_{\Omega_y})\big) 
            g(\phi_{\Omega_y}) 
        \right)
    \Bigg)\, d\phi
\end{align}

Now we need to finish the computation of the derivative of the derivative of self-attention component. 

Now we'll find
\begin{equation}
 \frac{\delta}{\delta g({\Phi})} \left[ g(\phi_{\Omega_x}) + \sum_{y} \text{softmax}_y\left(g(\phi_{\Omega_x}), g(\phi_{\Omega_y})\right) g(\phi_{\Omega_y}) \right]
\end{equation}

\begin{equation}
    = \frac{\delta}{\delta g({\Phi})} g(\phi_{\Omega_x}) + \sum_{y} \frac{\delta}{\delta g({\Phi})} \left[ \text{softmax}_y\left(g(\phi_{\Omega_x}), g(\phi_{\Omega_y})\right) \cdot g(\phi_{\Omega_y}) \right]
\end{equation}

\begin{equation}
    = \delta(\phi_{\Omega_x} - {\Phi}) + \sum_{y} \frac{\delta}{\delta g({\Phi})} \frac{e^{\langle g(\phi_{\Omega_x}), g(\phi_{\Omega_y}) \rangle}}{\sum_{y'} e^{\langle g(\phi_{\Omega_x}), g(\phi_{\Omega_{y'}}) \rangle}} \cdot g(\phi_{\Omega_y})
\end{equation}
where the first term is the Dirac delta function since the derivative is only non-zero on the patch of interest.

\noindent Now we need to use the product rule.

\begin{equation}
    = \delta(\phi_{\Omega_x} - {\Phi}) + \sum_{y} \left( 
    \frac{\delta}{\delta  g({\Phi})} \left( 
        \frac{e^{\langle g(\phi_{\Omega_x}), g(\phi_{\Omega_y}) \rangle}}{\sum_{y'} e^{\langle g(\phi_{\Omega_x}), g(\phi_{\Omega_{y'}}) \rangle}} 
    \right) \cdot g(\phi_{\Omega_y}) 
    + \left( 
        \frac{e^{\langle g(\phi_{\Omega_x}), g(\phi_{\Omega_y}) \rangle}}{\sum_{y'} e^{\langle g(\phi_{\Omega_x}), g(\phi_{\Omega_{y'}}) \rangle}} 
    \right) \cdot \frac{\delta}{\delta  g({\Phi})} g(\phi_{\Omega_y}) 
\right)
\end{equation}

From above, we know that the $\frac{\delta}{\delta  g({\Phi})} g(\phi_{\Omega_y})$ should just turn into a Dirac delta.

\begin{equation}
    = \delta(\phi_{\Omega_x} - {\Phi}) + \sum_{y} \left( 
    \frac{\delta}{\delta  g({\Phi})} \left( 
        \frac{e^{\langle g(\phi_{\Omega_x}), g(\phi_{\Omega_y}) \rangle}}{\sum_{y'} e^{\langle g(\phi_{\Omega_x}), g(\phi_{\Omega_{y'}}) \rangle}} 
    \right) \cdot g(\phi_{\Omega_y}) 
    + \left( 
        \frac{e^{\langle g(\phi_{\Omega_x}), g(\phi_{\Omega_y}) \rangle}}{\sum_{y'} e^{\langle g(\phi_{\Omega_x}), g(\phi_{\Omega_{y'}}) \rangle}} 
    \right) \cdot \delta(\phi_{\Omega_y} -\Phi) 
\right)
\end{equation}

Now all that remains is to use the chain rule on the first term. We're now looking just at this term here where we need to take the derivative of the attention value with respect to $g(\Phi)$. This attention value is a scalar and we're taking the derivative with respect to a vector. We'll denote the attention $\alpha_{xy}$

\begin{equation}
    \frac{\delta \alpha_{xy}}{\delta  g({\Phi})}  = \nabla_{g(\phi_{\Omega_x})} \alpha_{xy} \cdot \frac{\delta g(\phi_{\Omega_x})}{\delta g(\Phi)} + \nabla_{g(\phi_{\Omega_y})} \alpha_{xy} \cdot \frac{\delta g(\phi_{\Omega_y})}{\delta g(\Phi)} 
\end{equation}

Notice that the second term in both of these turn into Dirac delta functions. 

\begin{equation}
    \frac{\delta \alpha_{xy}}{\delta  g({\Phi})}  = \nabla_{g(\phi_{\Omega_x})} \alpha_{xy} \delta(\phi_{\Omega_x} - \Phi) + \nabla_{g(\phi_{\Omega_y})} \alpha_{xy} \delta(\phi_{\Omega_y} - \Phi) 
\end{equation} 

Now, for the derivative of the attention itself, recall that $\alpha_{xy} = \frac{e^{\langle g(\phi_{\Omega_x}), g(\phi_{\Omega_y}) \rangle}}{\sum_{y'} e^{\langle g(\phi_{\Omega_x}), g(\phi_{\Omega_{y'}}) \rangle}}$

Then we can compute
\begin{equation}
     \nabla_{g(\phi_{\Omega_x})} \alpha_{xy} = \nabla_{g(\phi_{\Omega_x})} \frac{e^{\langle g(\phi_{\Omega_x}), g(\phi_{\Omega_y}) \rangle}}{\sum_{y'} e^{\langle g(\phi_{\Omega_x}), g(\phi_{\Omega_{y'}}) \rangle}} = \alpha_{xy} \left( g(\phi_{\Omega_{y}}) - \sum_{y'} \alpha_{xy'} g(\phi_{\Omega_{y'}})\right)
\end{equation}

This is a standard attention derivative, so putting it all together we have

\begin{align}
0 
=\;
&\sum_x 
\int \pi_t(\phi)\, 
2\,
\Bigl(
    g(\phi_{\Omega_x}) 
    \;+\; 
    \sum_{y}
        \alpha_{xy}(\phi)\,g(\phi_{\Omega_y}) 
    \;-\;
    s[\phi](x)
\Bigr)^{T}
\nonumber\\
&\quad
\Biggl[
    % (1) Direct derivative wrt g(phi_{Omega_x})
    \underbrace{\delta\bigl(\phi_{\Omega_x}-\Phi\bigr)}_{
      \substack{\text{derivative wrt}\\g(\phi_{\Omega_x})}
    }
    \;+\;
    \sum_{y}
    \biggl(
       % (2) Chain rule wrt g(phi_{Omega_x}) in alpha_{xy}
        \underbrace{
            \alpha_{xy}(\phi)\,
            \Bigl[
                g(\phi_{\Omega_y}) 
                \;-\;
                \sum_{y'} \alpha_{xy'}(\phi)\,g(\phi_{\Omega_{y'}})
            \Bigr]
            \,\delta\bigl(\phi_{\Omega_x}-\Phi\bigr)
        }_{
          \substack{\text{chain rule wrt}\\g(\phi_{\Omega_x})\text{ in }\alpha_{xy}}
        }
        \nonumber\\
&\qquad\quad
        % (3) Chain rule wrt g(phi_{Omega_y}) in alpha_{xy}
        +\;
        \underbrace{
            \alpha_{xy}(\phi)\,
            \Bigl[
                g(\phi_{\Omega_x}) 
                \;-\;
                \sum_{y'} \alpha_{xy'}(\phi)\,g(\phi_{\Omega_{y'}})
            \Bigr]
            \,\delta\bigl(\phi_{\Omega_y}-\Phi\bigr)
        }_{
          \substack{\text{chain rule wrt}\\g(\phi_{\Omega_y})\text{ in }\alpha_{xy}}
        }
        %(4) Direct derivative wrt g(phi_{Omega_y})
        \;+\;
        \underbrace{
            \alpha_{xy}(\phi)\,\delta\bigl(\phi_{\Omega_y}-\Phi\bigr)
        }_{
          \substack{\text{derivative wrt}\\g(\phi_{\Omega_y})\text{ itself}}
        }
    \biggr)
\Biggr]
d\phi
\end{align}

We can simplify this. We claim that 
$$\sum_{y}
    \biggl(
       % (2) Chain rule wrt g(phi_{Omega_x}) in alpha_{xy}
        \underbrace{
            \alpha_{xy}(\phi)\,
            \Bigl[
                g(\phi_{\Omega_y}) 
                \;-\;
                \sum_{y'} \alpha_{xy'}(\phi)\,g(\phi_{\Omega_{y'}})
            \Bigr]
            \,\delta\bigl(\phi_{\Omega_x}-\Phi\bigr)
        }_{
          \substack{\text{chain rule wrt}\\g(\phi_{\Omega_x})\text{ in }\alpha_{xy}}
        } \bigg) = 0$$

\noindent Since setting $ \mu_x = \sum_{y'} \alpha_{xy'}(\phi)\,g(\phi_{\Omega_{y'}})$ and recalling that $\sum_y \alpha_{xy} =1$, these terms cancel. Finally, we recover the form

\begin{align}
0 =\ &  \sum_x \int \pi_t \left[ g(\phi_x) + \sum_y \alpha_{xy} g(\phi_{\Omega_y}) - s[\phi](x) \right]^T 
\delta(\phi_{\Omega_x} - \Phi)\ d\phi \nonumber \\
& + \int \pi_t \left[ g(\phi_x) + \sum_y \alpha_{xy} g(\phi_{\Omega_y}) - s[\phi](x) \right]^T 
\sum_y \alpha_{xy} \left( I + (g(\phi_{\Omega_x}) - \mu_x) \delta(\phi_{\Omega_y} - \Phi) \right) d\phi
% 0 =\ & 2 \sum_x \int \pi_t \left[ \tilde{g}[\phi](x) - s[\phi](x) \right]^T 
% \delta(\phi_{\Omega_x} - \Phi)\ d\phi \nonumber \\
% & + \int \pi_t \left[ \tilde{g}[\phi](x) - s[\phi](x) \right]^T 
% \sum_y \alpha_{xy} \left( I + (g(\phi_{\Omega_x}) - \mu_x) \delta(\phi_{\Omega_y} - \Phi) \right) d\phi
% \label{keyeqn}
\end{align}

\subsection{Top 1 Attention Derivation}
We begin with the same setup as above, where we assume that our attention is a ``winner-take-all'' regime, meaning that only the most attended to patch contributes to the sum. In particular, we have 
\[
      \sum_{y}\alpha_{xy}\,g(\phi_{\Omega_y})
      \;\longrightarrow\;
      g\bigl(\phi_{\Omega_{y^*(x)}}\bigr),
      \quad
      y^*(x)=\arg\max_y\langle g(\phi_{\Omega_x}),\,g(\phi_{\Omega_y})\rangle.
    \]

\noindent We also assume ``patch-independence'' under the distribution $\pi_t$ for all $t$, so that conditioning on $\phi_{\Omega_x} = \Phi$ does not change the distribution of $\phi_{\Omega_{y}}$ for $y \neq x$ and that our embedding $g$ is mean‐centered over patches (ie, \(\mathbb{E}_{\phi\sim\pi_t}\bigl[g(\phi_{\Omega_x})\bigr]=0 \quad\forall\,x\)).  We then substitute the top-1 attention form where $\sum_{y}\alpha_{xy}\,g(\phi_{\Omega_y}) = g\bigl(\phi_{\Omega_{y^*(x)}}\bigr)$ into  Equation ~\ref{keyeqn}. When we expand the the second term in Equation ~\ref{keyeqn}, we find that since we've assumed that $g$ is mean-centered and that we have approximate patch independence, the second term goes away, leaving 
\begin{equation}
\begin{aligned}
% 0=\;&
% \sum_{x}\!\int\!\pi_t(\phi)\;
%    2\bigl(g_x+g_{y^{*}(x)}-s[\phi](x)\bigr)^{\!\top}
%    \delta(\phi_{\Omega_x}-\Phi)\,d\phi \\[4pt]
% &+\;\sum_{x}\!\int\!\pi_t(\phi)\;
%    2\bigl(g_x+g_{y^{*}(x)}-s[\phi](x)\bigr)^{\!\top}
%    \delta(\phi_{\Omega_{y^{*}(x)}}-\Phi)\,d\phi .
0=\;&
\sum_{x}\!\int\!\pi_t(\phi)\;
   2\bigl(\tilde{g}[\phi](x)-s[\phi](x)\bigr)^{\!\top}
   \delta(\phi_{\Omega_x}-\Phi)\,d\phi \\[4pt]
&+\;\sum_{x}\!\int\!\pi_t(\phi)\;
   2\bigl(\tilde{g}[\phi](x)-s[\phi](x)\bigr)^{\!\top}
   \delta(\phi_{\Omega_{y^{*}(x)}}-\Phi)\,d\phi .
\end{aligned}
\end{equation}
Since there is a deterministic mapping between a given patch $\phi_{\Omega_x}$ and its most attended patch $\phi_{\Omega_{y^*(x)}}$, we see that the integrals with the delta peaks give closed form solutions:
\begin{equation}
    \int \tilde{g}[\phi](x)\delta(\phi_{\Omega_x} - \Phi) d\phi = \int \tilde{g}[\phi](x)\delta(\phi_{\Omega_y^*(x)} - \Phi) d\phi = g[\Phi] + g[\Phi^*] := \tilde{g}[\Phi],
\end{equation}
where $g[\Phi^*]$ denotes the output of the model on the patch most attended to by $\Phi$.

Thus, distributing the deltas, integrating over these terms, and moving them to the other side:
\begin{equation}
\begin{aligned}
& \tilde{g}(\Phi)\sum_{x}\pi_t(\phi_{\Omega_x} = \Phi) + \tilde{g}(\Phi)\sum_{x}\pi_t(\phi_{\Omega_{y^*(x)}} 
= \Phi) = \\ &\;
\sum_{x}\!\int\!\pi_t(\phi)s[\phi](x)
   \delta(\phi_{\Omega_x}-\Phi)\,d\phi
+\;
\sum_{x}\!\int\!\pi_t(\phi)s[\phi](x)
   \delta(\phi_{\Omega_{y^{*}(x)}}-\Phi)\,d\phi
\end{aligned}
\end{equation}
By the linearity of the integrals and sums we get:
\begin{equation}
\begin{aligned}
 \tilde{g}(\Phi)\sum_{x}\bigl[\pi_t(\phi_{\Omega_x} = \Phi) + \pi_t(\phi_{\Omega_{y^*(x)}} 
= \Phi)\bigr]  & = \;
\sum_{x}\!\int\!\pi_t(\phi)s[\phi](x)\bigl[
   \delta(\phi_{\Omega_x}-\Phi) +
   \delta(\phi_{\Omega_{y^{*}(x)}}-\Phi)\bigr]\,d\phi \\
   & = \sum_{x}\!\int\!\nabla_{\phi(x)}\pi_t(\phi)\bigl[
   \delta(\phi_{\Omega_x}-\Phi) +
   \delta(\phi_{\Omega_{y^{*}(x)}}-\Phi)\bigr]\,d\phi \\
   & = \sum_{x}\!\nabla_{\Phi(0)}\bigl[\pi_t(\phi_{\Omega_x} = \Phi ) + \pi_t(\phi_{\Omega_{y^{*}(x)}}=\Phi)\bigr]
\end{aligned}
\end{equation}
Dividing by $\bigl[\pi_t(\phi_{\Omega_x} = \Phi) + \pi_t(\phi_{\Omega_{y^*(x)}} 
= \Phi)\bigr]$, yields

\begin{equation}
\begin{aligned}
 \tilde{g}(\Phi) = \nabla_{\Phi(0)}\log\sum_{x}\!\bigl[\pi_t(\phi_{\Omega_x} = \Phi ) + \pi_t(\phi_{\Omega_{y^{*}(x)}}=\Phi)\bigr]
 \end{aligned}
 \end{equation}

\pagebreak
\section{Additional Results and Training Details}
\label{apndb}
\noindent We also trained a simple CNN with a top-1 attention layer on the end to better match our theory and a simple CNN with attention given by the simple identity matrix query and key matrices. Both of these models had a standard DDPM setup and we implemented our score network (or, equivalently noise predictor) as a very simple 2 layer CNN with 2 convolutional layers, where the first convolutional layer is a 2x2 convolution with stride 2 and hidden dimension 32, and our second is a transpose convolution also with a 2x2 kernel and stride length of 2. The top-1 attention layer used Gumbel Softmax since argmax (which would be the natural way to implement top-1 attention isn't differentiable \cite{jang2017categoricalreparameterizationgumbelsoftmax}. Because of this, training was more difficult, so we trained this model for $10,000$ epochs. Other than that, the training was the same as the two models mentioned above. We used a batch size of 64 over 5000 epochs, AdamW with learning rate $10^{-3}$ and weight decay $10^{-5}$ under a OneCycleLR schedule. Additionally, we maintain an exponential moving average (EMA) of the model parameters with decay 0.9999, updating it after each optimizer step. During sample generation, we use the EMA weights to improve stability \cite{kingma2015adam}. We also used a linear noise schedule and weight the MSE loss by $1 - \alpha_{\text{cumulative product}}[t]$, as is standard.\\

\noindent We found that both the CNN+Top1 Attention and CNN+Identity Attention both outperformed the CNN only architecture, but failed to achieve the same quantitative and qualitative results as the full attention model. In particular, Gumbel-Softmax-implemented Top-1 attrition resulted in $21.64 \%$ consistency across 100 trials, while the identity attention model resulted in $25.44 \%$ consistency across 100 trials.

\pagebreak

\section{Additional Samples}

\begin{figure}[b]
    \centering
    \includegraphics[width=1.0\linewidth]{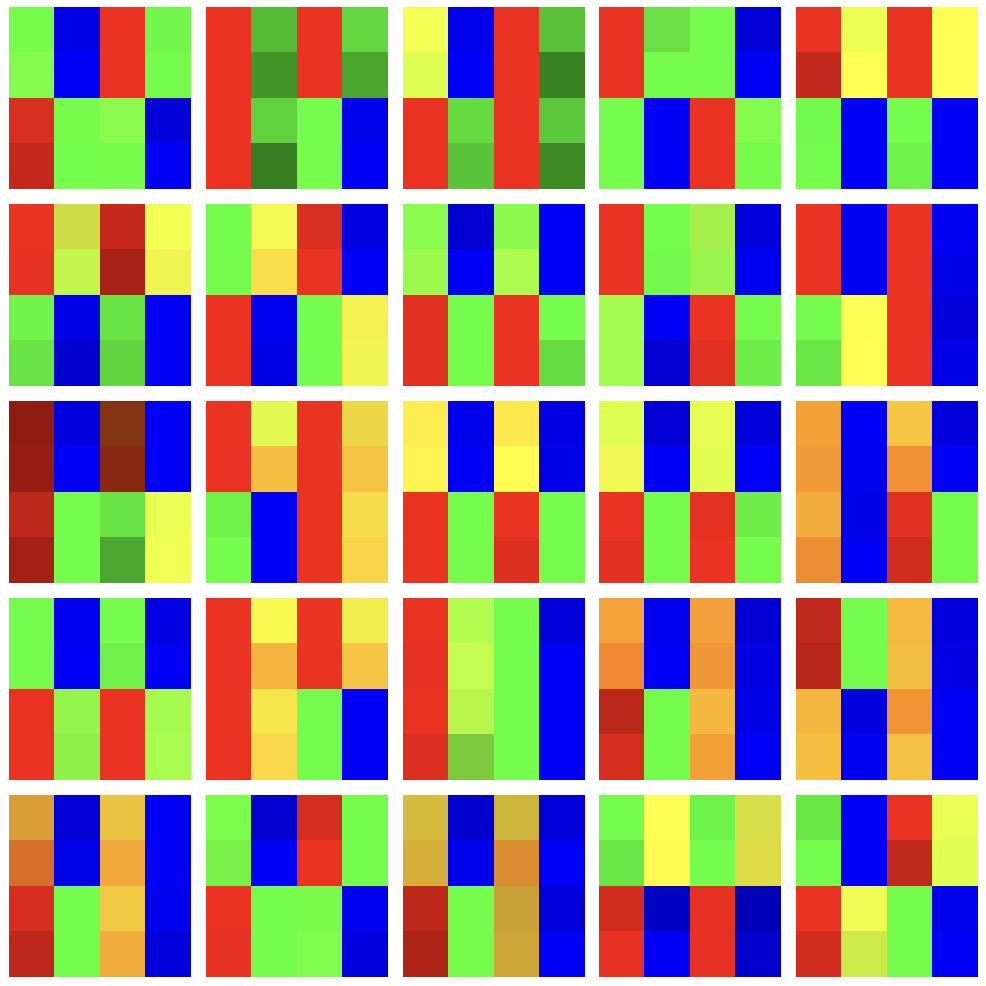}
    \caption{More Samples from CNN+Attention}
    \label{fig:attnsamples}
\end{figure}

\begin{figure}
    \centering
    \includegraphics[width=1.0\linewidth]{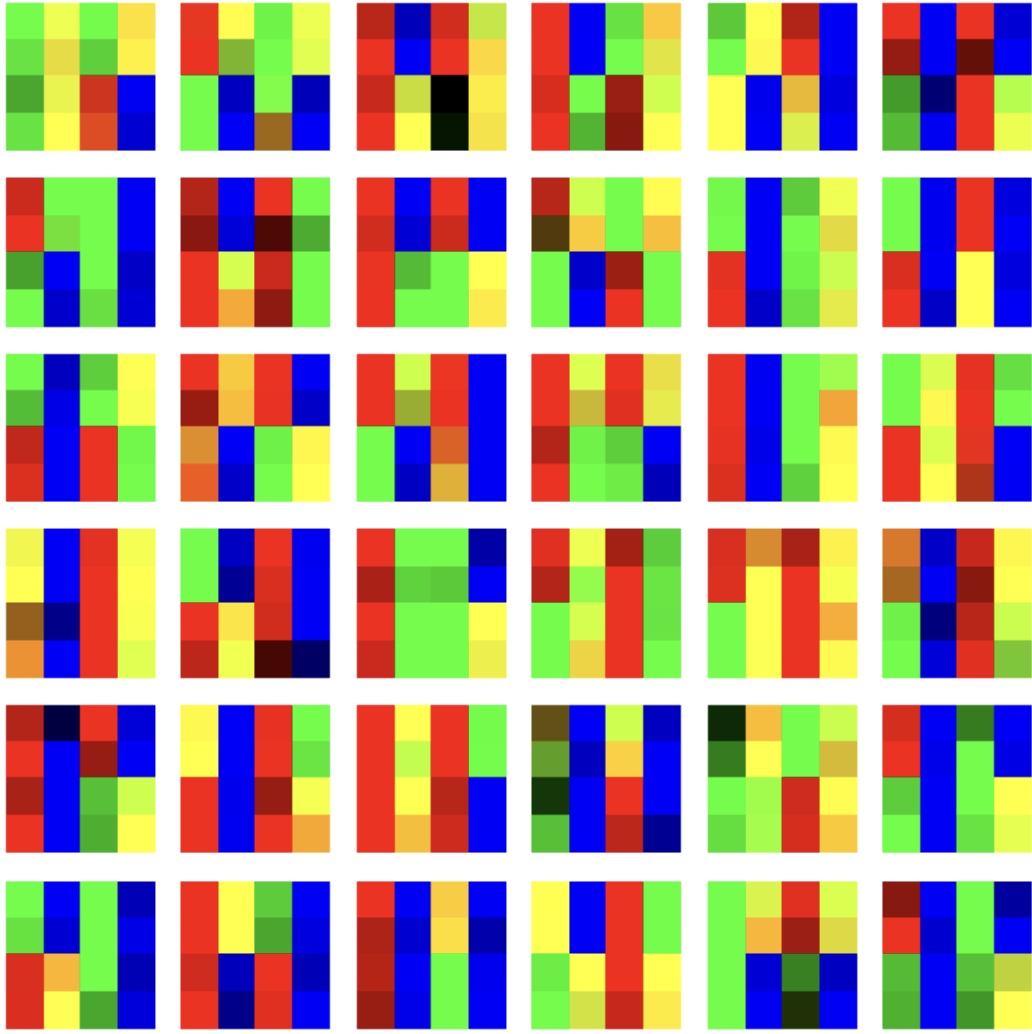}
    \caption{More Samples from CNN+ Top 1 Attention}
    \label{fig:top1attnsamples}
\end{figure}

\begin{figure}
    \centering
    \includegraphics[width=1.0\linewidth]{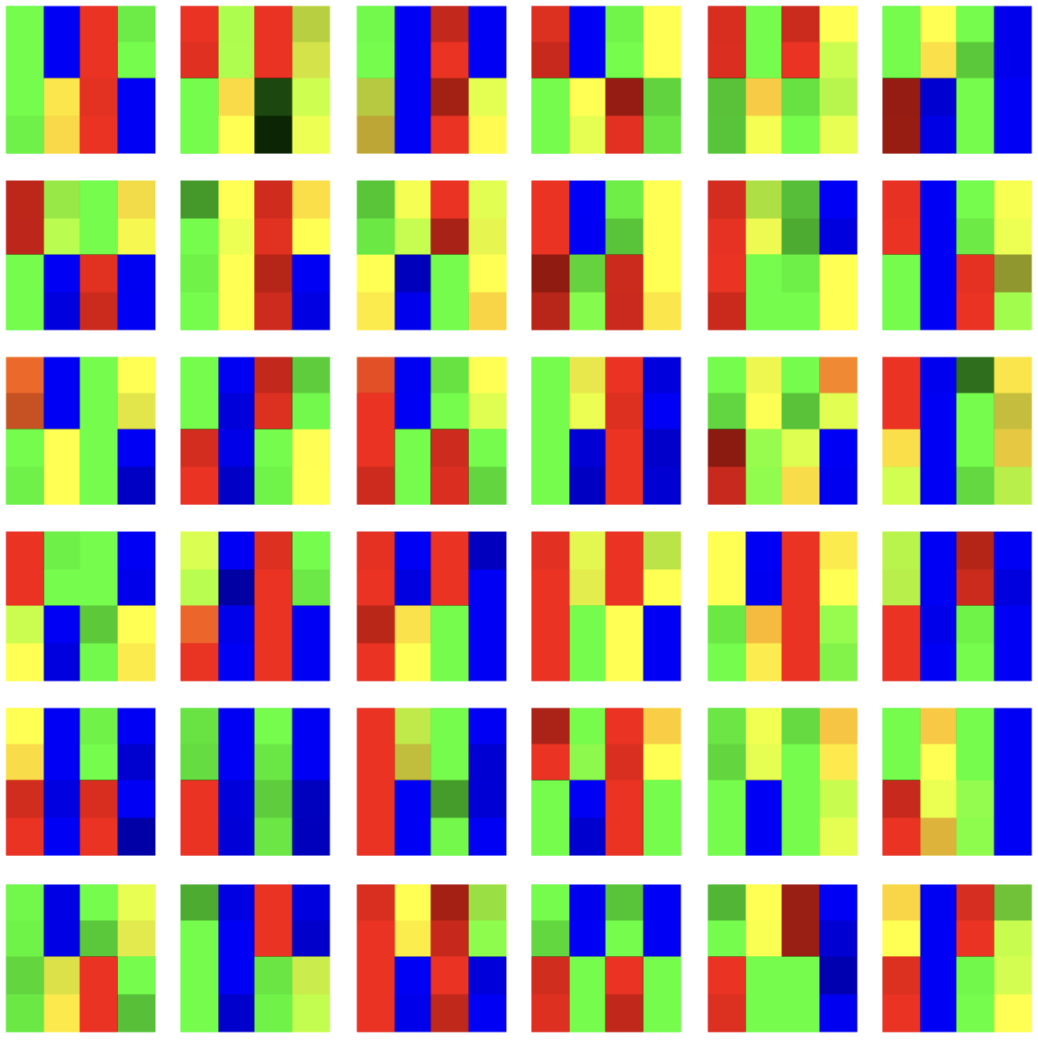}
    \caption{More Samples from Identity Attention}
    \label{fig:identityattention}
\end{figure}

\begin{figure}
        \centering
        \includegraphics[width=1.0\linewidth]{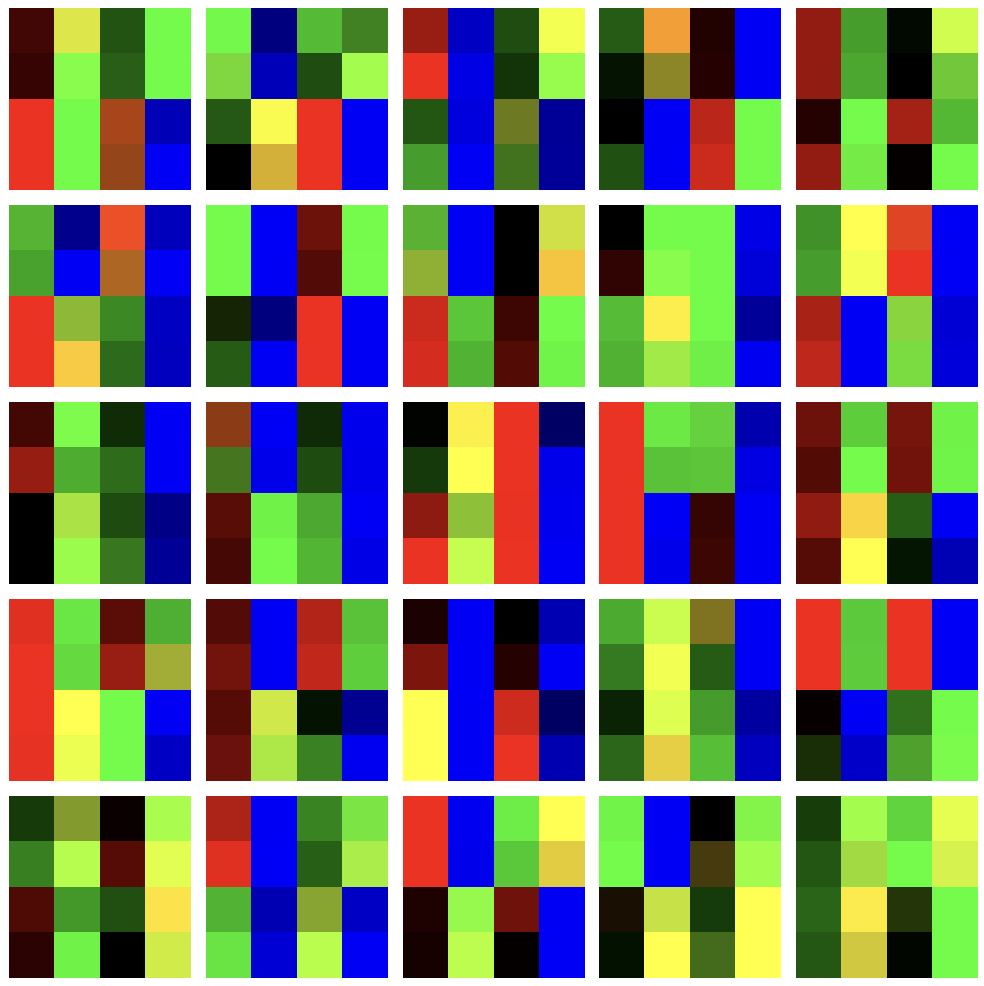}
        \caption{More Samples from CNN}
        \label{fig:cnnsamples}
\end{figure}

\end{appendix}

\end{document}